\documentclass[lettersize,journal]{IEEEtran}
\usepackage{amsmath,amsfonts}
\usepackage{algorithmic}
\usepackage{algorithm}
\usepackage{array}
\usepackage[caption=false,font=normalsize,labelfont=sf,textfont=sf]{subfig}
\usepackage{textcomp}
\usepackage{stfloats}
\usepackage{url}
\usepackage{verbatim}
\usepackage{graphicx}
\usepackage{cite}
\usepackage{listings}
\usepackage{hyperref}
\usepackage{xcolor}
\usepackage{multirow,tabularx}
\hypersetup{
  colorlinks=true,
}

\begin{document}

\title{PokerKit: A Comprehensive Python Library for Fine-Grained Multi-Variant Poker Game Simulations}

\author{Juho Kim \linebreak Faculty of Applied Science and Engineering \linebreak University of Toronto, Toronto, Ontario, Canada \linebreak juho.kim@mail.utoronto.ca}

\maketitle

\begin{abstract}
PokerKit is an open-source Python library designed to overcome the restrictions of existing poker game simulation and hand evaluation tools, which typically support only a handful of poker variants and lack flexibility in game state control. In contrast, PokerKit significantly expands this scope by supporting an extensive array of poker variants and it provides a flexible architecture for users to define their custom games. This paper details the design and implementation of PokerKit, including its intuitive programmatic API, multi-variant game support, and a unified hand evaluation suite across different hand types. The flexibility of PokerKit allows for applications in diverse areas, such as poker AI development, tool creation, and online poker casino implementation. PokerKit's reliability has been established through static type checking, extensive doctests, and unit tests, achieving 99\% code coverage. The introduction of PokerKit represents a significant contribution to the field of computer poker, fostering future research and advanced AI development for a wide variety of poker games. The source code is available at \url{https://github.com/uoftcprg/pokerkit}
\end{abstract}


\section{Introduction}
\IEEEPARstart{P}{oker}, a game that intertwines strategy and chance, has emerged as an exciting and challenging domain for artificial intelligence (AI) research, primarily due to its inherent nature of imperfect information. Pioneering AI agents such as Deepstack and Pluribus have leveraged poker, more specifically Texas hold 'em, as a benchmark for evaluating their algorithms \cite{doi:10.1126/science.aam6960, doi:10.1126/science.aay2400}. Nonetheless, poker is not a monolithic game but a collection of numerous variants, each introducing its unique set of complexities. This vast diversity underscores the inherent challenge of developing a comprehensive poker game logic system. This is particularly true in the open-source space, where solutions are often confined to supporting a limited set of poker variants \cite{pokerholdemengine, pypokerengine}.

In the open-source space, there exist numerous poker libraries. Some specialize in providing a simulated poker environment, but their focus tends to be on heads-up (2-player) no-limit Texas Hold’em \cite{pokerholdemengine, pypokerengine}. However, the vast majority of these libraries concentrate purely on hand evaluation and lack the extensive game logic needed for poker game simulations \cite{deuces, treys}.

To address this gap in the open-source space of computer poker, we developed PokerKit, a comprehensive Python library. It accommodates a multitude of poker variants, including but not limited to Texas hold 'em, Omaha hold 'em, seven card stud, razz, and deuce-to-seven triple draw. Moreover, PokerKit provides a programmatic interface that allows users to define their own variants and customize game parameters such as betting structures, high/low-split options, antes, blinds, and straddles.

This paper explores the development of PokerKit, delving into its features and capabilities, its rigorous testing and validation processes, and its existing and potential applications. Through this comprehensive examination, we aim to underscore PokerKit's unique contribution to the field of computational poker.

\section{Related Work}
The functionalities of PokerKit primarily fall into two categories: game simulations and hand evaluations. Game simulations encompass creating an environment where poker games can be played out programmatically, simulating real-world scenarios with high fidelity. On the other hand, hand evaluations are concerned with determining the strength of particular poker hands. In this section, we explore notable contributions in these areas, discussing their limitations and illustrating how PokerKit improves upon these existing solutions.

\subsection{Game Simulation}
Poker game logic has been developed for use in numerous online gaming platforms such as PokerStars, GGPoker, and 888Poker. While these implementations effectively serve their intended purposes, they are proprietary and usually support only a select few variants. Among these, PokerStars distinguishes itself by offering a broader range of game options \cite{pokerstars, ggpoker, 888poker}. Table \ref{tab:variants} provides an overview of the selections of games offered by these platforms.

PokerKit, in contrast, supports all major poker variants played on these platforms and much more, at any deck type, betting structure, limit, and optional high/low-split. This wide selection that surpasses the state-of-the-art online poker platforms further illustrates the comprehensive nature of PokerKit.

On the open-source front, libraries such as poker-holdem-engine and PyPokerEngine have made strides but have considerable limitations. They are limited in the number of variants they support and offer less intuitive interfaces. Some rely on networking, while others necessitate the usage of callbacks that return an action based on the game state \cite{pokerholdemengine, pypokerengine}. This methodology makes it impossible to carry out fine-grained control over game states, such as bet collection after each street, mandatory card turn-over to prevent chip dumping during all-ins, showdown order, choice of mucking the best hand (and vice versa), and killing losing hands post-showdown.

\begingroup
\setlength{\tabcolsep}{5pt}
\begin{table}[!t]
\caption{Supported game variants among online poker casinos \cite{pokerstars, ggpoker, 888poker}. Note that ``HL8'' denotes ``high/low-split eight or better'' and ``HLR'' denotes ``high/low-split regular''. The listing here contains all the variants defined in the 2023 World Series of Poker Tournament Rules \cite{2023wsoptourneys}. \label{tab:variants}}
\centering
\begin{tabular}{|c|c|c|c|c|}
\hline
Variants & PokerKit & PokerStars & GGPoker & 888Poker \\
\hline\hline
Texas hold 'em & yes & yes & yes & yes \\
\hline
Short-deck hold 'em & yes & yes & yes & no \\
\hline
Omaha & yes & yes & yes & yes \\
\hline
Omaha HL8 & yes & yes & no & yes \\
\hline
5-Card Omaha HL8 & yes & yes & no & no \\
\hline
7-card stud & yes & yes & no & no \\
\hline
7-card stud HL8 & yes & yes & no & no \\
\hline
7-card stud HLR & yes & no & no & no \\
\hline
Razz & yes & yes & no & no \\
\hline
A-to-5 triple draw & yes & no & no & no \\
\hline
2-to-7 triple draw & yes & yes & no & no \\
\hline
Badugi & yes & yes & no & no \\
\hline
Badacey & yes & no & no & no \\
\hline
Badeucey & yes & no & no & no \\
\hline
2-to-7 single draw & yes & yes & no & no \\
\hline
5-card draw & yes & yes & no & no \\
\hline
\end{tabular}
\end{table}
\endgroup

Comparatively, the Python chess engine, python-chess, offers an exemplary model. It allows users to interact programmatically with games through function calls \cite{pythonchess}. PokerKit emulates this style by providing an intuitive programmatic API to verify, query, and apply various operations.

\subsection{Hand Evaluation}
While it is not the primary focus of the library, PokerKit also provides facilities for evaluating poker hands. These tools support a greater number of hand types than those offered by alternative open-source libraries such as Deuces and Treys \cite{deuces, treys}, as shown in Table \ref{tab:hands}. Just like variants for game simulation, the end user may extend the existing framework to add their own hand type for their custom variants.

\begin{table}[!t]
\caption{Supported hand types among open-source libraries \cite{deuces, treys}. \label{tab:hands}}
\centering
\begin{tabular}{|c|c|c|c|}
\hline
Hands & PokerKit & Deuces & Treys \\
\hline\hline
Standard (Texas hold 'em, etc.) & yes & yes & yes \\
\hline
Eight-or-better & yes & no & no \\
\hline
Short-deck hold 'em & yes & no & no \\
\hline
Regular (Razz, etc.) & yes & no & no \\
\hline
Badugi & yes & no & no \\
\hline
Kuhn poker & yes & no & no \\
\hline
\end{tabular}
\end{table}

\section{Features and Capabilities}
\subsection{Game Simulation}
Each poker variant often introduces unique game rules and hand types not seen in other variants \cite{2023wsoptourneys}. The versatility of PokerKit allows for the customization of poker variants, allowing users to define their own unique games, adapt an existing variant, or implement a variant not currently supported by PokerKit out of the box. This flexibility is achieved without compromising the robustness of the implementation, backed by extensive unit tests and doctests to ensure error-free operations. Naturally, common variants come pre-defined in the PokerKit package, so, for most use cases, users will not have to define their own variants. An example of user-defined variant is shown in Figure \ref{fig:variant-definition}.

\begin{figure}[htbp]
\centering
\begin{lstlisting}
from pokerkit import (
    Automation as A,
    BettingStructure,
    Deck,
    KuhnPokerHand,
    Opening,
    State,
    Street,
)

state = State(
    # Automations,
    (
        A.ANTE_POSTING,
        A.BET_COLLECTION,
        A.BLIND_OR_STRADDLE_POSTING,
        A.CARD_BURNING,
        A.HOLE_DEALING,
        A.BOARD_DEALING,
        A.HOLE_CARDS_SHOWING_OR_MUCKING,
        A.HAND_KILLING,
        A.CHIPS_PUSHING,
        A.CHIPS_PULLING,
    ),
    Deck.KUHN_POKER,  # Deck
    (KuhnPokerHand,),  # Hand types
    # Streets
    (
        Street(
            False,  # Card burning
            (False,),  # Hole dealings
            0,  #  Board dealings
            False,  # Draw cards?
            Opening.POSITION,  # Opener
            1,  # Min bet
            # Max number of completions,
            # bets, or raises
            None,
        ),
    ),
    # Betting structure
    BettingStructure.FIXED_LIMIT,
    True,  # Uniform antes?
    (1,) * 2,  # Antes
    (0,) * 2,  # Blinds or straddles
    0,  # Bring-in
    (2,) * 2,  # Starting stacks
    2,  # Number of players
)
\end{lstlisting}
\caption{An initialization of a Kuhn poker game as a user-defined variant. Note that split-pot games can be created by specifying multiple hand types: one high and one low.}
\label{fig:variant-definition}
\end{figure}

PokerKit stands out with its ability to cater to an assortment of use cases, offering varying degrees of control over the game state. For instance, in use cases for poker AI agent development, where the agents' focus lies primarily in action selection during betting rounds, minute details such as showdown order, posting blinds, posting antes, and bet collection may not be pertinent. On the other hand, an online poker casino requires granular control over each game aspect. These include dealing hole cards one by one, burning cards, deciding to muck or show during the showdown (even when unnecessary), killing hands after the showdown, pushing chips to the winner's stack, and even the winner collecting the chips into their stack. PokerKit rises to this challenge, providing users with varying levels of automation tailored to their specific needs.

\subsubsection{The First Televised Million-dollar All-in Pot}
To demonstrate PokerKit, The first televised million-dollar all-in pot in poker history, played between professional poker players Tom Dwan and Phil Ivey, is recreated in Figure \ref{fig:simulation} \cite{dwanivey}. Note that, for the sake of conciseness in the following examples, operations like ante/blind posting, bet collection, showdown, hand killing, chip pushing, and chip pulling were configured to be automated. However, users have the option to manually invoke these operations by disabling the corresponding automation.

\begin{figure}[htbp]
\centering
\begin{lstlisting}
from pokerkit import (
    Automation as A,
    NoLimitTexasHoldem,
)

state = NoLimitTexasHoldem.create_state(
    # Automations
    (
        A.ANTE_POSTING,
        A.BET_COLLECTION,
        A.BLIND_OR_STRADDLE_POSTING,
        A.CARD_BURNING,
        A.HOLE_CARDS_SHOWING_OR_MUCKING,
        A.HAND_KILLING,
        A.CHIPS_PUSHING,
        A.CHIPS_PULLING,
    ),
    True,  # Uniform antes?
    500,  # Antes
    (1000, 2000),  # Blinds or straddles
    2000,  # Min-bet
    # Starting stacks
    (1125600, 2000000, 553500),
    3,  # Number of players
)
# Pre-flop
state.deal_hole("Ac2d")  # Ivey
# Antonius (hole cards unknown)
state.deal_hole("5h7s")
state.deal_hole("7h6h")  # Dwan
# Dwan
state.complete_bet_or_raise_to(7000)
# Ivey
state.complete_bet_or_raise_to(23000)
state.fold()  # Antonius
state.check_or_call()  # Dwan
# Flop
state.deal_board("Jc3d5c")
# Ivey
state.complete_bet_or_raise_to(35000)
state.check_or_call()  # Dwan
# Turn
state.deal_board("4h")
# Ivey
state.complete_bet_or_raise_to(90000)
# Dwan
state.complete_bet_or_raise_to(232600)
# Ivey
state.complete_bet_or_raise_to(1067100)
state.check_or_call()  # Dwan
# River
state.deal_board("Jh")
# Final stacks: 572100, 1997500, 1109500
print(state.stacks)
\end{lstlisting}
\caption{An example game simulation. Note that the game was created as a pre-defined variant.}
\label{fig:simulation}
\end{figure}

\subsection{Hand Evaluation}

Beyond its use of providing a simulated poker environment, PokerKit serves as a valuable resource for evaluating poker hands. The hand evaluation component in PokerKit offers a high-level Pythonic programmatic interface for hand evaluation, as demonstrated in Figure \ref{fig:evaluation}. It supports the largest selection of hand types in any mainstream open-source poker library, and can easily be extended to support custom hand types. This makes it an invaluable tool for users interested in studying the statistical properties of poker, regardless of their interest in game simulations.

\begin{figure}[htbp]
\centering
\begin{lstlisting}
from pokerkit import (
    Card,
    OmahaHoldemHand,
)

h0 = OmahaHoldemHand.from_game(
    "6c7c8c9c",  # Hole cards
    "8s9sTc",  # Board cards
)
h1 = OmahaHoldemHand("6c7c8s9sTc")

print(h0 == h1)  # True
\end{lstlisting}
\caption{An example hand evaluation.}
\label{fig:evaluation}
\end{figure}

\section{Design and Implementation}
\subsection{Game Simulation}
PokerKit’s poker simulations are architected around the concept of states, encapsulating all the vital information about the current game through its member variables.

\begin{list}{-}{}
\item{Cards in deck.}
\item{Community cards.}
\item{Mucked cards.}
\item{Burned cards.}
\item{Bets}
\item{Stacks}
\item{Hole cards}
\item{Pots (main + side)}
\item{And more...}
\end{list}

\begin{figure}[ht]
\centering
\includegraphics[width=2in, clip, keepaspectratio]{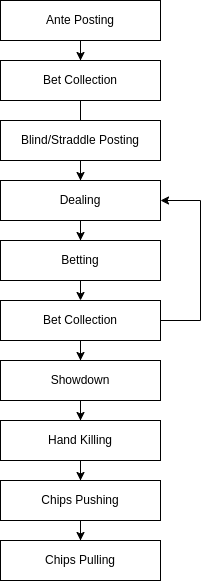}
\caption{The flowchart of phases in PokerKit.}
\label{fig:phases}
\end{figure}

PokerKit structures the game flow into distinct phases, each supporting a different set of operations. The flowchart of phases is shown in Figure \ref{fig:phases}. Depending on the game state, each phase may be skipped. For instance, if the user has specified no antes, the ante posting phase will be omitted. Likewise, if no bets were placed during the betting phase, the bet collection phase will be bypassed. A phase transition occurs upon the completion of a phase. This transition is internally managed by the game framework, facilitating a seamless game flow to the end user.

During each phase of PokerKit's game simulation, the user can invoke various methods to execute operations. Each operation belongs to a specific phase and can only be enacted when the corresponding phase is active.

\subsubsection{Ante Posting}
During the ante posting phase, each player has the option to execute an ante-posting operation. The parameters supplied to the state during its creation may dictate no antes, uniform antes, or non-uniform antes, such as big blind antes. If no player is due to post an ante, this phase is bypassed.

\subsubsection{Bet Collection}
The collection of bets on the table occurs after any phase that allows players to bet. If any bet is present, the bet collection operation must be performed before proceeding to the subsequent phase. This phase only occurs after ante posting or betting. When no bets are pending collection, this phase is skipped.

\subsubsection{Blind or Straddle Posting}
Forced bets like blinds or straddles must be posted before the start of the first street. PokerKit accommodates a variety of blind or straddle configurations, ranging from small and big blinds, to button blinds, or even no blind at all. If the state is configured to exclude any forced bets, this phase is skipped.

\subsubsection{Dealing}
The dealing phase precedes a betting phase. During this phase, the user can deal board or hole cards, contingent upon the state's configuration. Options to burn a card or discard and draw cards are also available when applicable. This phase is bypassed if only one player remains in the hand.

\subsubsection{Betting}
During betting, players can execute actions such as folding, checking, calling, posting a bring-in, completing, betting, or raising. During state creation, the user must specify how to select the first player to act and the betting limits. This phase is bypassed if all players are all-in or if only one player remains in the hand.

\subsubsection{Showdown}
During the showdown, players reveal or muck their hands in accordance with the showdown order. The first to show is typically the last aggressor in the final street. If no one bet, the player who was the first to act in the final betting round must show first. Players can opt to show a losing hand or muck a winning hand, even though this is often disadvantageous. When dealing with all-in pots, players are obligated to show their hands in order to prevent chip-dumping \cite{2023wsoptourneys}. If this is the case, or if only one player remains in the pot, the showdown phase is bypassed.

\subsubsection{Hand Killing}
The dealer is responsible for ``killing,'' or discarding, hands that cannot win any portion of the pot. If no hand should be killed, this phase is bypassed.

\subsubsection{Chips Pushing}
The dealer is charged with pushing the chips to the winners. In poker games, the pot size is always non-zero due to the mandatory presence of antes, forced bets, or bring-ins (as enforced by PokerKit). Thus, this phase is always carried out.

\subsubsection{Chips Pulling}
Players may incorporate the chips they've won back into their stack. In poker, at least one player is guaranteed to win the pot. Consequently, this phase is never skipped.

\begin{table}[!t]
\caption{State phases and their corresponding operations. \label{tab:pops}}
\centering
\begin{tabular}{|c|c|}
\hline
Phases & Operations \\
\hline\hline
Ante posting & Ante posting \\
\hline
Bet collection & Bet collection \\
\hline
Blind or straddle posting & Blind or straddle posting \\
\hline
\multirow{4}{*}{Dealing} & Card burning \\
\cline{2-2} & Hole dealing \\
\cline{2-2} & Board Dealing \\
\cline{2-2} & Standing pat or discarding \\
\hline
\multirow{4}{*}{Betting} & Folding \\
\cline{2-2} & Checking or calling \\
\cline{2-2} & Bring-in posting \\
\cline{2-2} & Completion, betting, or raising to \\
\hline
Showdown & Hole cards showing or mucking \\
\hline
Hand killing & Hand killing \\
\hline
Chips pushing & Chips pushing \\
\hline
Chips pulling & Chips pulling \\
\hline
\end{tabular}
\end{table}

\begin{table}[!t]
\caption{State operations and their automatabilities. \label{tab:aops}}
\centering
\begin{tabular}{|c|c|}
\hline
Phases & Automatable \\
\hline\hline
Ante posting & Yes \\
\hline
Bet collection & Yes \\
\hline
Blind or straddle posting & Yes \\
\hline
Card burning & Yes \\
\hline
Hole dealing & Yes \\
\hline
Board Dealing & Yes \\
\hline
Standing pat or discarding & No \\
\hline
Folding & No \\
\hline
Checking or calling & No \\
\hline
Bring-in posting & No \\
\hline
Completion, betting, or raising to & No \\
\hline
Hole cards showing or mucking & Yes \\
\hline
Hand killing & Yes \\
\hline
Chips pushing & Yes \\
\hline
Chips pulling & Yes \\
\hline
\end{tabular}
\end{table}

The operations in PokerKit for each phase and their automatabilities are summarized in Table \ref{tab:pops} and \ref{tab:aops}. Each operation is coupled with two associated methods: a verification method and an action query, as seen in Figure \ref{fig:opmethods}. The verification method validates if a move can be executed within the rules, considering the current game state and the variant in play. It raises an error if any discrepancy is detected. Users can directly invoke this or use a corresponding action query method (with optional arguments), which simply checks if the verification method triggers an error and returns a boolean value indicating the validity of the action. The method that performs the operation initially runs the verification method, executing the operation only if no errors are raised. If the verification fails, the state remains unchanged.

\begin{figure}[htbp]
\centering
\begin{lstlisting}
def verify_operation(self, ...):
    ...
    if is_inoperable:
        raise ValueError("...")
    ...

def can_operate(self, ...):
    try:
        self.verify_operation(...)
    except ValueError:
        return False
    return True

def operate(self, ...):
    self.verify_operation(...)
    ...
\end{lstlisting}
\caption{The method triplets of an example operation.}
\label{fig:opmethods}
\end{figure}

PokerKit's philosophy is that it should focus on maintaining the game state and enforcing rules. Error handling is left to the user, who may need to handle errors differently depending on the application. Assertions are used sparingly throughout the library for sanity checks. On runtime, users may choose to disable them for optimization purposes without impacting the library's functionalities.

The game variant is defined through several attributes, which are as follows:

\begin{list}{-}{}
\item{Deck: Most variants use a 52-card deck.}
\item{Hand types: Most variants have one, but high/low-split games have two.}
\item{Streets: Each specifies whether to burn a card, deal the board, deal the players, draw cards, the opener, the minimum bet, and the maximum number of bets or raises.}
\item{Betting structure: Betting limits such as no-limit, pot-limit, or fixed-limit.}
\end{list}

This flexibility in the definition gives PokerKit the ability to describe not only every variant specified in the 2023 World Series of Poker Tournament Rules \cite{2023wsoptourneys} but also esoteric variants, ranging from Greek hold 'em, Kuhn poker, 6-card Omaha, Rhode Island hold 'em, and countless more.

In addition to the parameters related to the variants, users can supply additional values, namely automations, antes (uniform antes or non-uniform antes such as big blind antes), blinds/straddles, bring-ins, and starting stacks.

\subsection{Hand Evaluation}
Hand evaluation is another vital aspect of PokerKit. The library generates a lookup table for each hand type. The hands are generated in the order or reverse order of strength and assigned indices, which are used to compare hands. High-level interfaces allow users to construct hands by passing in the necessary cards and using standard comparison operators to compare the hand strengths. It’s worth noting that “strength” in poker hands does not necessarily mean “low” or “high” hands \cite{2023wsopliveaction}. Each hand type in PokerKit handles this distinction internally, making it transparent to the end user.

\begin{table}[!t]
\caption{Hand types and their lookup tables in PokerKit \cite{treys}. \label{tab:tables}}
\centering
\begin{tabular}{|c|c|c|c|}
\hline
Hand types & Lookups & Table size \\
\hline
\hline
Standard high & \multirow{4}{*}{Standard} & \multirow{4}{*}{7462} \\
\cline{1-1}
Standard low & & \\
\cline{1-1}
Greek hold 'em & & \\
\cline{1-1}
Omaha hold 'em & & \\
\hline
Eight-or-better low & \multirow{2}{*}{Eight-or-better} & \multirow{2}{*}{112} \\
\cline{1-1}
Omaha eight-or-better low & & \\
\hline
Short-deck hold 'em & Short-deck hold 'em & 1404 \\
\hline
Regular low & Regular & 7462 \\
\hline
Badugi & Badugi & 1092 \\
\hline
Kuhn poker & Kuhn poker & 3 \\	
\hline
\end{tabular}
\end{table}

In the lookup process, cards are converted into unique integers that represent their ranks. Each rank corresponds to a unique prime number and the converted integers are multiplied together. The suitedness of the cards is then checked. Using the product and the suitedness, the library looks for the matching hand entries which are then used to compare hands. This perfect hashing approach was originally employed in the Deuces hand evaluation library and is also used by Treys \cite{deuces, treys}.

\section{Performance Benchmarks}
The benchmarks in this section were conducted using a computer equipped with a 12th Gen Intel\textregistered{} Core\texttrademark{} i7-1255U processor and 16 GB of RAM, utilizing Python version 3.11.5.

Note that the specific values reported here reflect the performance under this particular hardware and software setup and may vary on different systems.

\subsection{Game Simulation}
The general performance of the game simulation component of PokerKit has been evaluated by digitalizing all 83 televised hands from the final table of the 2023 World Series of Poker (WSOP) \$50,000 Poker Players Championship, shown in Table \ref{tab:benchmark-game-simulation}. The selection of this particular tournament was motivated by its varying number of players (as the finalists were eliminated), and its rotation through nine different diverse variants of poker \cite{2023wsopevent43}.

\begin{table}[!t]
	\caption{The game simulation speed of PokerKit. \label{tab:benchmark-game-simulation}}
\centering
\begin{tabular}{|c|c|}
\hline
& PokerKit \\
\hline
\hline
Speed (\# games/s) & 847.31 \\
\hline
\end{tabular}
\end{table}

\subsection{Hand Evaluation}
The hand evaluation facilities in PokerKit sacrifice slight speed to provide an intuitive high-level interface. As a result, PokerKit's hand evaluation suite is slower than open-source Python hand evaluation libraries such as Treys as shown in Table \ref{tab:speeds} \cite{treys}. Note that the Deuces library, another popular alternative, is not benchmarked as it lacks Python 3 support and Treys is a fork of Deuces that supports Python 3 \cite{deuces, treys}.

For the benchmarking process, we iterated through all possible Texas hold 'em hands in both PokerKit and Treys. The speeds are reported as the number of hands evaluated per second.

\begin{table}[!t]
\caption{The hand evaluation speeds of PokerKit and Treys \cite{treys}. \label{tab:speeds}}
\centering
\begin{tabular}{|c|c|c|c|}
\hline
& PokerKit & Treys \\
\hline
\hline
Speed (\# hands/s) & 1016740.7 & 3230966.4 \\
\hline
\end{tabular}
\end{table}

\section{Testing and Validation}
Ensuring robustness and error-free operation has been a top priority throughout PokerKit's development. To validate its functionality, a variety of rigorous testing methods were employed, spanning from extensive doctests to unit tests, and recreating renowned poker hands and games.

\subsection{Static Type Checking}
While Python is a dynamically typed language, it also provides an option to write type annotations in the code \cite{pep484}. Mypy, a static type checker for Python, is utilized to scrutinize the PokerKit project \cite{mypy}. All the code in PokerKit successfully passes the Mypy static type checking with the ``\--\--strict'' flag.

\subsection{Doctests}
Docstrings serve a dual purpose of providing not only documentation and sample usage but also tests \cite{pythondocsdoctest}. Most classes and methods within the library are accompanied by example usages in doctests that illustrate their functionality. Furthermore, these doctests offer insight into potential scenarios where errors are raised, helping users understand the library's expected behavior under various circumstances.

\subsection{Unit Tests}
Unit tests are employed to test broader functionalities that do not fall into normal usage patterns. For example, some of the unit tests evaluate all possible hand for each lookup and validates the hands by comparing them against the results from other poker hand evaluation libraries. For efficiency, this is achieved by obtaining an MD5 checksum of the large text string of ordered hands generated by the lookup tables in PokerKit and other open-source libraries.

One of the methods of validating PokerKit has been the recreation of various famous poker hands and games. This includes all 83 televised hands from the final table of the 2023 World Series of Poker (WSOP) \$50,000 Poker Players Championship which features poker games of nine different variants \cite{2023wsopevent43}. The library was tested for consistency, ensuring that pot values after each street and resulting stacks after the hand is over accurately match the real outcomes.

Additional unit tests were implemented to rigorously scrutinize PokerKit's behavior under unusual circumstances like scenarios involving extremely low starting stacks that often fall below the blinds or antes.

Together, the doctests and unit tests cover 99\% of the PokerKit codebase, ensuring a comprehensive examination.

In conclusion, through the diligent application of static type checking, doctests, unit tests, and recreations of real-life hands, PokerKit has undergone thorough testing and validation. This rigorous approach guarantees that the library can dependably handle a wide array of poker game scenarios, solidifying its position as an invaluable resource for researchers and developers in the field.

\section{Use Cases and Applications}
Thanks to its flexible architecture and fine-grained control, PokerKit is adaptable to a wide array of tasks. It proves to be an effective tool in various applications, ranging from AI development and poker tool creation to serving as the backbone of online poker platforms. This section explores key use cases where PokerKit has demonstrated its invaluable contributions.

\subsection{AI Development}
The continuous action space and the imperfect information nature inherent to poker present a compelling challenge for AI. Most of the breakthroughs in Poker AI agents were limited to Texas hold 'em variants \cite{doi:10.1126/science.aam6960, doi:10.1126/science.aay2400}. PokerKit's capacity to support a comprehensive set of poker variants, coupled with its robust and error-free nature, makes it an ideal framework for developing, testing, and benchmarking poker AI models that can generalize beyond Texas hold’em. Its design, which allows for the automation of irrelevant operations, facilitates the simulation of numerous game scenarios needed for effective AI training and evaluation. Furthermore, its benchmarked speed is adequate for various AI applications.

One popular domain of computer poker is the development of poker solvers. Generally, solvers like PIOSolver begin with a game tree construction, which is then followed by the more computationally intensive Nash equilibrium calculation \cite{upi}. PokerKit can be leveraged during this initial game tree construction while the Nash equilibrium calculation can be carried out in more performant languages such as C or C++. It's worth noting that during the game tree construction, the sequence of possible actions is independent of the cards dealt. This intrinsic characteristic provides ample opportunities for developers to implement optimizations and reduce the runtime by orders of magnitude.

\subsection{Poker Tool Development}
PokerKit lays a solid foundation for the development of various poker tools, including hand equity calculators and poker training software. With its extensive support for most major poker variants, developers are empowered to create tools that cater to a broad spectrum of poker games. Further, PokerKit's efficient hand evaluation capability and its intuitive programmatic API enable the development of user-friendly, performance-optimized tools.

\subsection{Online Poker Platform Development}
The application of PokerKit extends beyond research and tool development; it's equally effective in creating online poker platforms. Its capacity to simulate the intricate changes of the poker table makes it ideal for the smooth gameplay experience demanded in online poker rooms.

In summary, whether it's developing sophisticated AI models, creating intuitive poker tools, or providing a reliable basis for online poker platforms, PokerKit's versatile capabilities make it well-suited to a wide array of applications in the poker landscape.

\section{Conclusion}
This paper presents PokerKit, a versatile and efficient Python library for poker game simulation and hand evaluation. By providing support for various major poker variants, PokerKit ensures a comprehensive and accurate implementation of poker game logic and hand evaluation mechanisms.

A key strength of PokerKit is its intuitive design, which allows for both manual and automatic control over game states, facilitating its application in diverse contexts. The library’s built-in unit tests and doctests ensure the robustness of its implementation, while its use in real-world projects provides a testimony of its reliability.

Despite PokerKit's extensive functionality, there are still exciting prospects for future work in the world of computer poker. Notably, the absence of a standardized interface for Poker AI, similar to the Universal Chess Interface, presents an opportunity for development. While the PIOSolver team has created a Universal Poker Interface, it does not adequately represent the broad scope of poker or poker AI due to its specialized commands and parameters for its Nash equilibrium solving algorithm and its two-player limitation \cite{upi}. A standardized interface catering to a variety of AI methodologies could substantially enhance the efficiency and interoperability of poker AI development.

In conclusion, PokerKit presents a valuable contribution to the field of computer poker, offering a comprehensive and efficient toolkit for poker game simulation and hand evaluation. Future work, including the creation of a universal poker AI interface, holds the promise of further enriching the field.

\section{Acknowledgement}
We acknowledge the use of ChatGPT, based on GPT-4, an AI language model, in order to enhance the readability of the paper by correcting grammar, polishing language, and improving clarity throughout all sections of the paper \cite{openai2023gpt4}.

\bibliographystyle{IEEEtran}
\bibliography{references}

\vfill

\end{document}